\def\year{2020}\relax
\documentclass[letterpaper]{article} 
\usepackage{aaai20}  
\usepackage{times}  
\usepackage{helvet} 
\usepackage{courier}  
\usepackage[hyphens]{url}  
\usepackage{graphicx} 
\usepackage{graphicx}  
\usepackage{amsmath}

\title{Deep Time-Stream Framework for Click-Through Rate Prediction by Tracking Interest Evolution}
\author{Shu-Ting Shi$^{1,2}$, Wenhao Zheng$^2$, Jun Tang$^2$, Qing-Guo Chen$^2$, Yao Hu$^2$, Jianke Zhu$^3$, Ming Li$^1$\thanks{Corresponding Author} \\ $^1$National Key Laboratory for Novel Software Technology, Nanjing University, Nanjing 210023, China \\ $^2$YouKu Cognitive and Intelligent Lab, Alibaba Group, Hangzhou, China \\ $^3$ Zhejiang University, Hangzhou, China
\\ shist@lamda.nju.edu.cn, \{zwh149850, donald.tj, qingguo.cqg, yaoohu\}@alibaba-inc.com, jkzhu@zju.edu.cn, lim@nju.edu.cn}
 
\begin{document}
	
	\maketitle
	\begin{abstract}
		Click-through rate (CTR) prediction is an essential task in industrial applications such as video recommendation.
		Recently, deep learning models have been proposed to learn the representation of users' overall interests, while ignoring the fact that interests may dynamically change over time.
		We argue that it is necessary to consider the continuous-time information in CTR models to track user interest trend from rich historical behaviors.
		In this paper, we propose a novel Deep Time-Stream framework (DTS) which introduces the time information by an ordinary differential equations (ODE). 
		DTS continuously models the evolution of interests using a neural network, and thus is able to tackle the challenge of dynamically representing users' interests based on their historical behaviors.
		In addition, our framework can be seamlessly applied to any existing deep CTR models by leveraging the additional Time-Stream Module, while no changes are made to the original CTR models.
		Experiments on public dataset as well as real industry dataset with billions of samples demonstrate the effectiveness of proposed approaches, which achieve superior performance compared with existing methods.
	\end{abstract}
	
	\section{Introduction}
	Click-through rate (CTR) prediction aims to estimate the probability of a user clicking on a given item, which has drawn increasing attention in the communities of academia and industry.
	In the example of a video website, a CTR algorithm is deployed to provide users with videos from thousands of different categories, and thus it is crucial to precisely capture users' interests so that they will keep using the website longer and bring more revenue to the website.

	To accomplish this goal, the key problem is how to model user interest based on user historical clicks which reflects user preference.
	To extract representation of user's interests, many models have been proposed from traditional methodologies \cite{friedman2001greedy,rendle2010factorization} to deep CTR models \cite{guo2017deepfm,qu2016product,lian2018xdeepfm}.
	Although these models achieve great success in modeling users' overall interests, they are ignorant of the dynamic changes in users' preferences. 
	To pursue a more precise result, RNN-based methods \cite{wu2017recurrent,hidasi2016general,zhou2019deep} have been proposed to capture the dependencies in user-item interaction sequences.
	However, these methods only consider the order of users' behaviors and ignore the \textit{time interval} between behaviors which is the important information on predicting users' behaviors. As an example, in Figure 1, Mike usually watches videos about Donald Trump during the day while enjoys music videos of Taylor Swift at night, according to his behaviors' timestamps. Thus, regarding Mike's playlog only as a sequence of clicked videos would neglect changes of his latent interests over time. 
	Unfortunately, existing CTR models do not have the capability to model the pattern on the continuous-time, since most of them are unaware to the time interval.

	\begin{figure}[t]
		\centering
		\includegraphics[width=1.0\columnwidth]{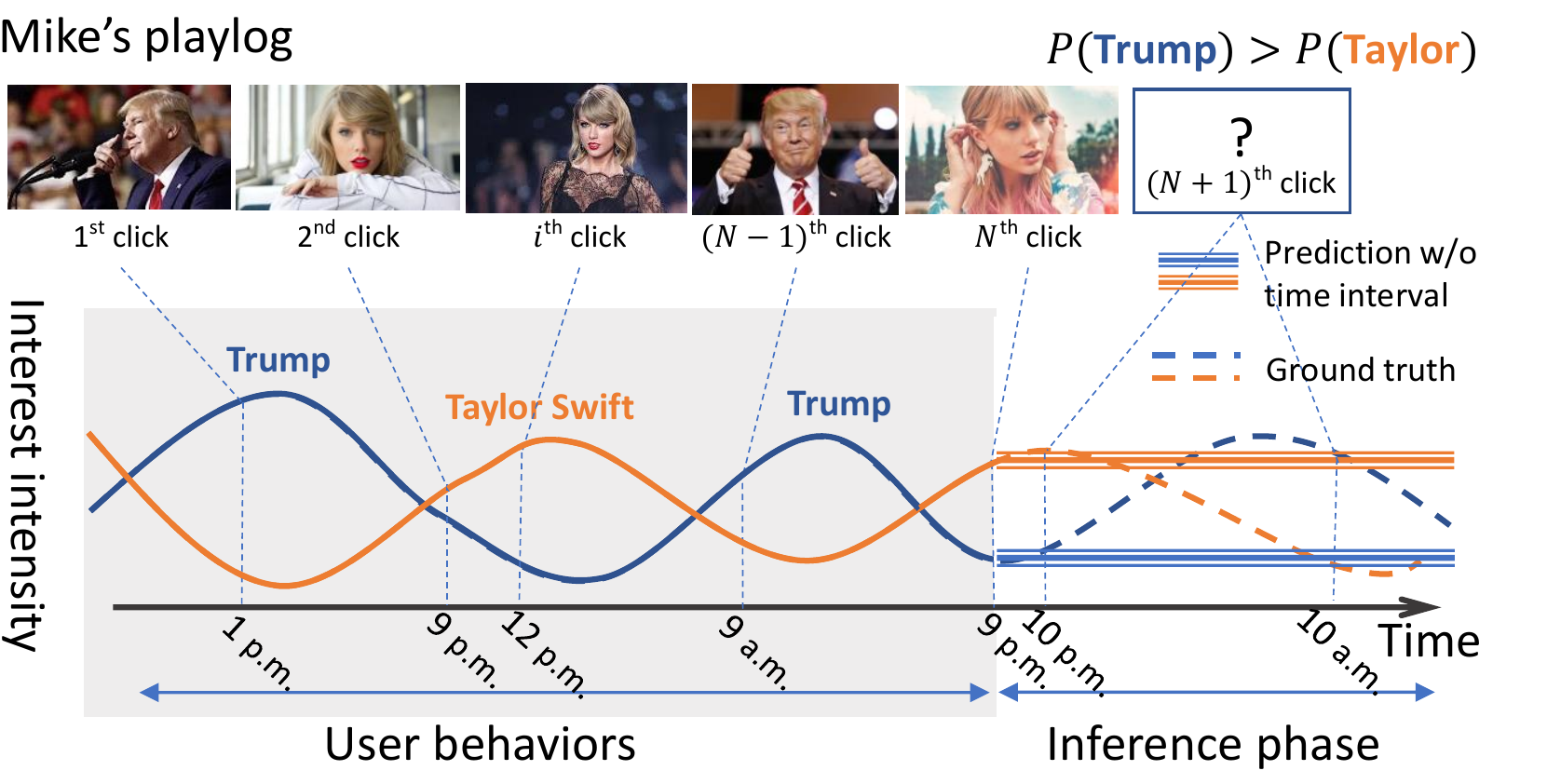} 
		\caption{
		The user interests evolve on the time-stream. Only when considering intervals among user behaviors could the sequential pattern be captured.}
		\label{fig1}
	\end{figure}
	
	Besides, at the inference phase, it is problematic that only predicting the \textit{next} click without considering \emph{when} the action will be performed.
	Incorporating the time of users' behaviors, \textit{i.e.} modeling the effects of elapsed time interval between behaviors, is important in accurately modeling users' interest. For example, in Figure 1, if Mike watches a video of Taylor at 9 p.m., it more likely that he will watch another video of Taylor than Donald in a few hours, while the probability of watching videos of Donald should be significantly higher after half a day. However, traditional methods always get the exactly same prediction at any time.

	Based on the aforementioned observations, we argue that it is crucial to consider time-stream information, \textit{i.e.} continuous-time information, in CTR models. 
	Therefore, we propose a novel Deep Time-Stream framework (DTS), which introduces the time-stream information into CTR model. Time-stream information could be formulated by ordinary differential equations (ODE), which refers to a function that describes the relationship between a dependent variable's derivative and the independent variable. 
	Specifically, DTS leverages ODE to model the evolution of users' latent interests, by parameterizing the derivative of users' latent interests states with respect to time, such that the solution of ODE describes the dynamic evolution of users' interests.
	Moreover, DTS is equipped with the ability to unify users' historical behaviors (what have clicked) and target items (what will click) on the time-stream by clicks' timestamp, thus would make inference corresponding to the given \textit{next time} and provide a more precises CTR prediction.
	To archieve the minimum model-altering cost, the ODE is packaged as a Time-Stream Module which can be applied in any deep CTR models. 
	The contributions of this paper are summarized as follows:
	\begin{itemize}
		\item We propose a novel DTS framework that models users' latent interests evolution as an ODE, which significantly improves the expressive ability of models and can better capture the evolving characteristics of users' interests.
		\item DTS can generate users' feature at an arbitrary time and thus allows flexible as well as adaptive evaluations.
		\item The Time-Stream Module can be easily transplanted into existing CTR models without changing the original structure.
	\end{itemize}

	\section{Background}
	\label{section:background}
	In machine learning, it is a crucial task to efficiently conduct a class of hypothesis, linear or nonlinear, that can represent the data patterns. 
	\textit{Ordinary Differential Equations (ODEs)} can also be used as a hypothesis.
	Considering the differential equation in $R^d$:
	$
	\frac{dz}{dt} = f(z,t), \ z(0)=z_0,
	$
	the solution of $z$ at time $t$ is denoted as $z(t)$. 
	The basic idea behind the ODE approaches to supervised learning is to tune $f$ so that the map $z(t)$ can produce nonlinear function needed to fit the data.
	
	In fact, \cite{chen2018neural} reveals that deep neural networks can be considered as discrete ODE, and their iterative
	updates can be regarded as an Euler discretization of a continuous transformation.
	On the other hand, neural ODEs are a family of deep neural network models that can be interpreted as a continuous equivalent of Residual Networks (ResNets) or Recurrent Neural Networks (RNNs). 
	To see this, consider the transformation of a hidden state from a
	layer $t$ to $t + 1$ in ResNets or RNNs:
	\begin{align}
	\label{Eq:RNN}
	h_{t+1} = h_t + f_t(h_t).
	\end{align}
	In ResNets, $h_t\in R^d$ is the hidden state at layer $t$ and $f_t: R^d \to R^d$ is some differentiable function which preserves the dimension of $h_t$.
	In RNNs, $h_t\in R^d$ is the hidden state at $t$-th RNN cell which update throw a function $f_t: R^d \to R^d$.
	The difference $h_{t+1}-h_t$ can be interpreted as a discretization of the derivative $h'(t)$ with timestep $\Delta t=1$. 
	Letting $\Delta t\to 0$, we see that the dynamically hidden state can be parameterized by an ODE:
	$
	\lim_{\Delta t\to 0} \frac{h_{t+\Delta t}-h_t }{\Delta t} = f(h, t).$

	This solution of $z(t)$ or $h(t)$ can be solved using an ODE solver with many sophisticated numerical methods to choose, e.g. linear multi-step methods, Runge-Kutta methods \cite{runge1895numerische,Kutta1901Beitrag} and adaptive time-stepping. Above methods could be helpful on deep learning, since they could adaptively choose the layers of network.
	It should be noted that here our concern is not the solver itself, but rather the representation of the data.
	So we refered to the solver as a black-box differential equation solver:
	\begin{align}
	\label{Eq:solver}
	z_{t_1},\cdots,z_{t_N} &= ODEsolve(z_{t_0}, f, \theta_f, t_1,\cdots,t_N)
	\end{align}
	where $\theta_f$ is the parameters of $f$.
	
	
	%
	
	In next section, we show how the ODEs are utilized to model the dynamics of users' interest evolution, and how to make ODEs stable while training.

	\section{The Deep Time-Stream Framework}
	\label{section:Framework}

	In this section, we describe the details of our proposed model. We firstly formalize CTR as a binary classification problem.
	Given a sample of data $x=(x^U, x^V, x^P)\in \mathcal{X}$, where $(x^U, x^V, x^P)$ denotes the collection of the concatenate of different fields' one-hot vectors from \textit{User behavior}, \textit{target Video} and \textit{user Profiles}, respectively.
	Moreover, each field contains a list of click behaviors, $x^U = [(v_{1}, c_{1});(v_{2}, c_{2});...;(v_{N}, c_{N})]$, in which $x^U_i=(v_i, c_i)$ denotes the video $v_i$ and corresponding category $c_i$ at the $i$-th behaviors that happens at time $t_i$, $N$ is the number of user's history behaviors;
	$x^V$ denotes the target video and its category $x^V=(v_{N+1},c_{N+1})$, and the equation is established because the target video should happen as the $(N+1)$-th user click, the predicting time of this potential click is refers to as \textit{next time} $t_{N+1}$. 
	Thus, we unify the user historical behaviors and target video on the time-stream by the their timestamps denoted as $t$, $t=[t_1, t_2,\cdots, t_N, t_{N+1}]$.
	\textit{User Profiles} $x^P$ contains useful profile information such as \textit{gender}, \textit{age} and so on.
	Label $y\in \mathcal{Y}$ indicates whether the user click the specific video, $y=1$ means click while $y=0$ means not.
	The goal of CTR is to learn a mapping $h \in \mathcal{H}$ from $\mathcal{X}$ to $\mathcal{Y}$, where $\mathcal{H}$ denotes the hypothesis space, $h: \mathcal{X}\mapsto \mathcal{Y}$ to predict whether the user will click the video. 
	The prediction function $h$ can be learned by minimizing the following objective function:
	\begin{align}
	\label{eq:objective}
	\min_{h}\sum_{(x,y)\in \mathcal{X}\times \mathcal{Y} } \mathcal{L}(h(x;t),y) 
	\end{align}
	where $\mathcal{L}$ is the empirical loss which will be introduced in detail in following subsections. 
	
	\subsection{General Framework}
	
\begin{figure}[t]
		\centering
		\includegraphics[width=1.0\columnwidth]{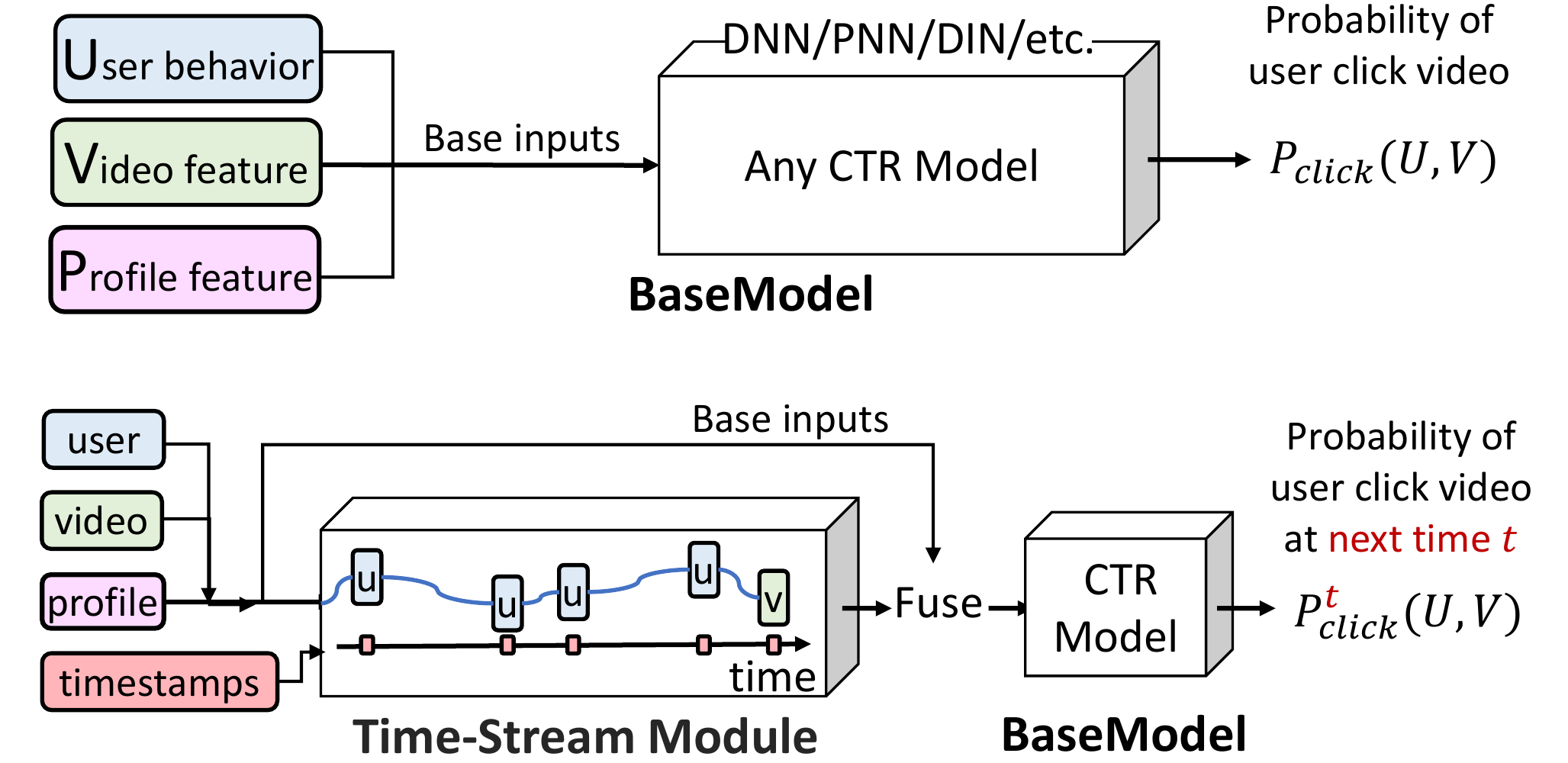}  
		\caption{DTS = BaseModel + Time-Stream Module. The Time-Stream Module introduces continuous-time information and by a fuse operation the base inputs are enhanced that can be feed into basemodel to get a more precise result.}
		\label{fig2}
	\end{figure}

	Our proposed framework DTS could be regarded as BaseModel plus Time-Stream Module, as shown in Figure~\ref{fig2}. BaseModel is referred to an existing deep CTR model such as DNN~\cite{covington2016deep}, PNN~\cite{qu2016product} and DIN~\cite{zhou2018deep}.
	Aside from the basemodel, Time-Stream Module collects the timestamps of all the events, including a user's historical click time in the past and the user's potential click time in the predicting moment. Note that the latter part are ignored in existing CTR models.
	Moreover, Time-Stream Module tracks the latent interest evolution by an ODE to calculate an enhanced input which introduces the continuous-time information while preserves the dimensions of base inputs.
	Thus, any deep CTR model can be used as the BaseModel in our DTS framework without changes made.
	Compared with the BaseModel that output an click probability on the event of user click item, DTS can improve the output by predicting the click probability on the event of user click item at given time.
	
	In the following subsections, we would explain the structure of BaseModel, and introduce the Time-Stream Module that are used for capturing interests and then modeling interest evolution.
	
	\subsection{BaseModel}
	\label{subsection:Review of BaseModel}
	Most deep CTR models are built on the basic structure of Embedding-Pooling-MLP. 
	The basic structure is composed of several parts:
	\begin{itemize}
		\item \textbf{Embedding} Embedding layer is the common operation that transforms the sparse feature into low-dimensional dense feature. 
		$x^U$, $x^V$ and $x^P$ are embedded as $(e_{1},...,e_{N})$, $e_{N+1}$ and $e^P$, respectively.
		In the field of \textit{User Profiles}, the sparse features embedded as $e^U$
		\item \textbf{Pooling} The embedding vectors are fed into pooling operation.
		$
		e^U = Pooling(e_{1},...,e_{N})
		$
		where $e^U$ refers to as user vector. The pooling can be sum pooling, average pooling, or specially designed attentive pooling\cite{zhou2018deep}.
		\item \textbf{Multilayer Perceptron (MLP)}  All these pooled vectors from different categories are concatenated. 
		Then, the concatenated vector is fed into a following MLP for final prediction.
	\end{itemize}
	
	\subsubsection{Target Loss}
	A widely used loss function in deep CTR models is negative log-likelihood function, which uses the label of target item to supervise overall prediction:
	\begin{align}
	\label{eq:targetloss}
	\mathcal{L}_{target} = -\frac{1}{N} \sum_{i=1}^N (y_i\log p(x_i) + (1-y_i)\log(1-p(x_i))) 
	\end{align}
	
	where $x_i = (x_{i}^U, x_{i}^V, x_{i}^P) \in \mathcal{D}$, $\mathcal{D}$ is the training set of size $N$.
	$y_i \in \{0, 1\}$ represents whether user clicks target item. 
	$p(x)$ is the output of network, which is the predicted probability that the user clicks target item.
	
	\subsection{Time-Stream Module}
	\label{subsection:Time-Stream}
	Users' interests are dynamic over time rather than static.
	BaseModel obtains a representation vector of user interest by a pooling operation over the clicked item feature but ignores the time information.
	The absence of dynamic pattern restricts the power of user behavior feature, which plays a key role in modeling user interests since the user clicked items are the expression of a user's interest at the corresponding time. 
	For BaseModel, the lack of ability for modeling continuous pattern leads to the inaccuracy to model the dynamic user interest.
	\begin{figure*}[ht]
		\centering
		\includegraphics[width=1.0\textwidth]{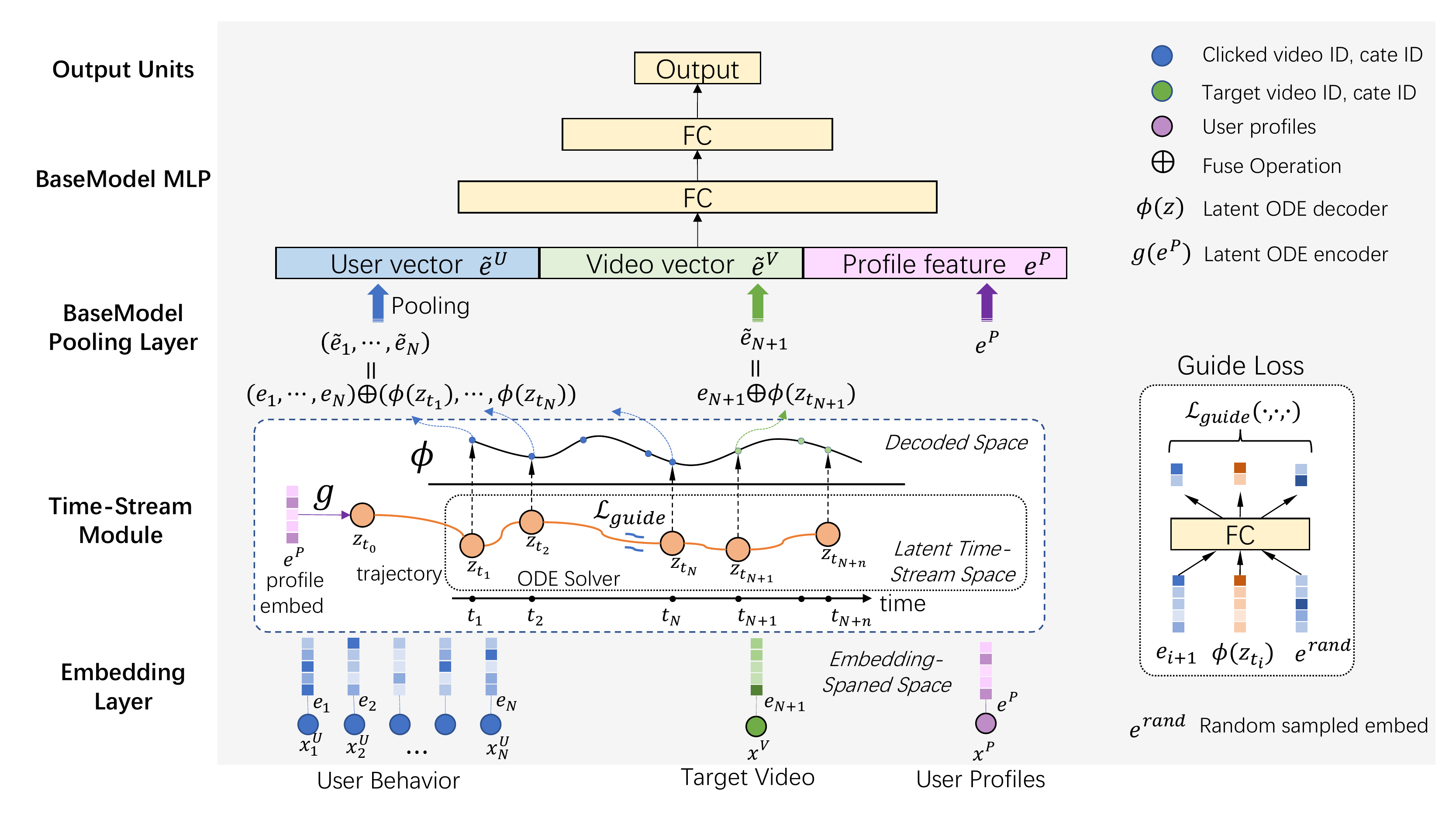}  
		\caption{The structure of Time-Sream Module. DTS keeps the basic framework of BaseModel thus inheriting their proven performance. Moreover, DTS extends Time-Sream module that the latent time state $z_t$ is modeled as an ODE. Decoder $\phi$ mapping $z_t$ to embedded space fused with embedding to enhance the quality of embedding. Guide loss is designed to help the hidden state converge}
		\label{fig3}
	\end{figure*}
	
	Is there an elegant way to represent a user's real-time interests and model the dynamic interest evolution pattern?
	The nature of continuous-time evolving inspires us to design a novel method named Time-Stream Framework that leveraging ODE to model the dynamic interest trace.
	ODE have been applied  to a wide variety of fields such as physics, biology, chemistry, engineering and economics and if the ODE can be solved, given an initial point it is possible to determine all its future positions, a collection of points known as a trajectory or orbit.
	In this paper we novelly using ODEs as hypothesis class that the trajectory denotes a latent interest evolution trace.
	As mentioned in Eq~\ref{Eq:RNN}, ODE can be a general form of RNNs and RNNs can be thought of as a discretization of the continuous ODE.
	There are several advantages with continuous ODE approach such as the flexible evaluations, which corresponds to choosing the RNN lengths adaptively.
	Moreover, we can also use advanced numerical methods for training, such as the multi-grid method or the parallel shooting method.
	Figure 3 illustrates the architecture of the Time-Stream Module.

	In details, to represent the interest evolution by a latent trajectory of ODE, a differentiable function $f$ is used, $\frac{dz(t)}{dt} = f(z(t),t;\theta_f)$ denotes the interest evolution rate, where $\theta_f$ is the parameters of $f$.
	Thus, given a initial state $z_{t_0}$ the trajectory of ODE can be solved using a solver mentioned in Equation~\ref{Eq:solver}:
	\begin{align}
	\begin{split}
	z_{t_1},\cdots,&z_{t_N},z_{t_{N+1}} \\
	&= ODEsolve(z_{t_0}, f, \theta_f, t_1,\cdots,t_N,t_{N+1}),   
	\end{split}
	\end{align}
	in which, where $z_{t_1},\cdots, z_{t_N},z_{t_{N+1}}$ is the solution of ODE which describes the latent state of dynamics $f$ at each observation time $t_1,\cdots,t_N,t_{N+1}$. 
	Since similar people are more likely to have similar interest evolution pattern, we construct a mapping $g$ that 
	transforms the user profile embedding $e^P$ to the latent time-stream space to obtain the initial value:
	$
	z_{t_0} = g(e^P;\theta_g),
	$
	the mapping $g$ is a linear transformations with parameter $\theta_g$ and serves as an encoder from profile embedding space to the latent time-stream space.

	On the other hand, $\phi$ is the decoder that transform latent time-stream feature $z_{t_i}$ to the video embedding-spaned space.
	$\phi(z_{t_i};\theta_\phi)$ is the adjustment or supplementary of behavior feature which carries additional behavioral evolving patterns.
	For the adjustment of user behavior feature, we have:
	$
	\widetilde{e}_{i} = e_{i} + \phi(z_{t_i};\theta_\phi).
	$
	where $i=1,2,...,N$. The fuse operation can be set as other operation such as concatenation, but in this work the add operation are used which keeps the adjustment and original feature equally contribute.
	For the target video feature, we have
	$
	\widetilde{e}^{V} = e_{N+1} + \phi(z_{t_{N+1}};\theta_\phi).
	$
	
	
	The enriched behavior feature $\widetilde{e}^U = (\widetilde{e}_{1},\widetilde{e}_2,...,\widetilde{e}_N)$, video vector $\widetilde{e}^V$ and profile feature $e^P$ was then send into the rest part of Base CTR model.

	Using ODEs as a generative model allows us to make predictions for arbitrary time, whether in the past or in the future since the on timeline are continuous.
	The output of ODE can be computed by a black-box differential equation solver, which evaluates the hidden unit dynamics wherever necessary to determine the solution with the desired accuracy.
	
	\subsubsection{The choice of function $f$}
	The latent function $f$ needs to be specified and different types of functions can be used to meet different requirements.
	Next, we will introduce several approaches that leverage different type of ODE function $f$ to model the process of interest evolution.
	\begin{itemize}
		\item \textbf{Simple form.} This is the simplest form of function $f$ under the assumption that $f$ is the function of independent variable $t$:
		\begin{align}
		f(z,t) =\frac{dz}{dt}=A(t), \ z(t) = \int^t_{t_0} A(\lambda)d\lambda + C,
		\end{align}
		where $A$ is a control function, and $C$ is a constant that can be solved given a initial state. 
		This type of problem may have an analytical solution that by computing $z(t)$ directly. If so, there is no extra cost needed to solve the ODE numerically.
		A special case is linear differential equation with constant coefficients $f(z,t)=A(t)=\alpha$ which means the latent state discount at rate $\alpha$. 
		Thus $z_{t_i}=\alpha (t_i -t_0) + z_{t_0} $ holds for all $t$.
		The insight is that the monotonous trajectory of $f$ mimic the characteristic of user interest: is mainly influenced by recent interests, so one should reduce the impact of early interest and increase the impact of the user’s recent behavior.
		This special case is extremely simple but achieves surprising performance shown in experiment.
		
		\item \textbf{Complex form.} The aforementioned simple form of $f$ cannot express user's diverse time-series pattern. 
		To overcome this limitation, another choice is to parameterize the derivative of the dynamics $f$ using a neural network which improves the expressive ability of model greatly. 
		In this paper, a two layer neural network with sigmoid activation unit is used:
$
		f \left( z \right) =\sigma(w_2 \cdot \sigma(w_1 \cdot z +b_1) +b_2),
		$

where $w_1,w_2,b_1,b_2$ are the linear parameters and $\sigma(.)$ is the activate unit.
		It is hard to obtain an analytical solution in this form of $f$. 
		The solution on $z_{t_1},\cdots,z_{t_N},z_{t_{N+1}}$ is computed using a numerical ODE solver mentioned in the Background. 
		
	\end{itemize}
	
	\subsubsection{Guide Loss}
	The aforementioned functions can be solved on a single call to ODE toolbox and modern ODE solvers provide guarantees on the growth of approximation error.
	However, we have several concerns:
	1) When the form of function becoming complicated, the behavior of ODE may encounter situations of explodes, converges to stationary states or exhibits chaotic behavior.
	This may explain some of the difficulties, e.g., the vanishing and explosion of gradients encountered in the training of deep neural networks.
	2) On the other hand, since the click behavior of target item is triggered by users' interest evolution, the label only indicates the last of click behavior $z_{t_{N+1}}$, while history state $z_t$ can not obtain proper supervision.

	To alleviate these problems, we propose \textit{guide loss}, which uses behavior embedding $e_{i}$ to supervise the learning of latent functions.
	To do this, inspired by the loss of Word2Vec~\cite{mikolov2013distributed}, we build a small network that push the decoded hidden state $\phi(z_{t_i})$ more close to the next behavior $e_{i+1}$ than a random negative sampled instance $e^{rand}$.
	Guide loss can be formulated as:
	\begin{align*}
	\mathcal{L}_{guide}(p,v,n) = -\frac{1}{N}\sum_{i}(v_i \cdot p_i+v_i \cdot n_i -\log(\frac{v_i \cdot p_i}{v_i \cdot n_i})),\\
	p_i = FC(e_{i+1}), \  v_i = FC(\phi(z_{t_i})), 	\ n_i = FC(e^{rand}).
	\end{align*}
where $FC(x)$ is a fully connected layer with PRelu as activation.
	The overall loss  in our model is:
	\begin{align}
	\mathcal{L} = \mathcal{L}_{target} + \lambda \mathcal{L}_{guide} 
	\end{align}
	where $\mathcal{L}$ is the overall loss function, $\mathcal{L}_{target}$ is introduced in Eqution \ref{eq:targetloss} and $\lambda$ is the hyper-parameter which balances the interest representation and CTR prediction.

	Overall, the introduction of g{}uide loss has several advantages: 
	1) from the aspect of interest learning, the introduction of guide loss helps each hidden state of ODE represent interest expressively. 
	2) As for the optimization of ODE, guide loss reduces the difficulty of backpropagation when ODE models long history behavior sequence.
	3) Guide loss gives more semantic information for the learning of the embedding layer, which leads to a better embedding matrix.
	
    \subsubsection{Training and inference}
	At the training phase, our model is equipped with the ability to reload the parameters of the BaseModel. Then all the weights are finetuned to get a quick convergence.
	We would achieve a \textit{safe-start} by initializing the parameters of $f$ and initial value to zeros, such that the trajectory of ODE is a constant of zero.
	Thus, at the start of training, the overall model stay the same as the original CTR base model.
	
	At the inference phase, we could predict the user interest evolution at the arbitrary recommendation time $t_{N+1}$,  since we leverage ODE solver to integrate the function $f$ at next time $t_{N+1}$.
	In industrial, DTS would be efficient: When predicting multiple CTR at $t_{N+1},t_{N+2}$ and $t_{N+n}$, there is no need to compute the hidden trajectory from scratch. It is easy to integrate the function $f$ from $t_{N}$ to $t_{N+n}$ that is cheap for computation. 
	

	\section{Experiments}
	\label{section:experiments}
	To evaluate the effectiveness of the Deep Time-Stream framework, we conduct experiments on public datasets and industrial dataset.
	In these experiments, the proposed Time-Stream Module is applied on multiple BaseModels.
	Note that DTS is a framwork that inheriting the base CTR model then extending the time-stream module.
	Therefore the effectiveness of DTS should be reflected in the improvement compared with BaseModel. 
	The comparison metrics will be introduced below.

	\subsection{Datasets and Experimental Settings}
	We use both public and industrial datasets to verify the effect of the Time-Stream Module.
	\subsubsection{Public Dataset}
	Amazon Dataset contains product reviews and metadata from Amazon, which is used as benchmark dataset in many works \cite{he2016ups,mcauley2015image}. 
	We conduct experiments on a subset named \textit{Electronic}, which contains 192,403 users, 63,001 goods, 801 categories and 1,689,188 samples.
	For \textit{User Behavior}, features include user-reviewed goods\_id list, category\_id list, and corresponding timestamp list.
	For \textit{Target Item}, features include target goods\_id, category\_id and the next time that denote when this click prediction is made.
	Dataset collects user behaviors that happens at time $t_1,t_2,\cdots,t_N$, where $N$ is the number of users history behaviors. DTS can naturally handle different N,  which corresponds to choosing the RNN lengths adaptively. Since some baselines contain RNN, for all dataset, the max N is set as 100 to keep the comparison fair.
	The task is to predict the $(k+1)$-th reviewed goods by making use of the first $k$ reviewed goods. 
	Training dataset is generated with $k = 1, 2,\cdots$ $,N-2$ for each user.
	In the test set, we predict the $N$-th good given the first $N-1$ reviewed goods.
	\subsubsection{Industrial Dataset}
	The industrial dataset is constructed by user playlog and profile information from a video platform.
	Similar to the public dataset, we collect features including video\_id, cate\_id, user-watched video\_id list and cate\_id list.
	Overall, 1.7 billion samples has been collect including 1.4 million users, 6.3 million videos, 278604 categories.
	For \textit{Profile features}, user profiles such as gender, age, activity score are used.
	Training dataset is generated with $k = 1, 2,\cdots,$ $N-2$ for each user.
	In the test set, we predict the $N$-th video given the first $N-1$ watched videos.
	\subsubsection{Compared Methods}
	We set BaseModels as some mainstream CTR prediction methods to evaluate the effectiveness of the Time-Stream framework. 
	The BaseModels are used as:
	\begin{itemize}
		\item \textbf{DNN} \cite{covington2016deep} DNN takes the setting of Embedding\&Pooling\&MLP and sum pooling operation was used to integrate behavior embeddings.
		\item \textbf{Wide\&Deep} \cite{cheng2016wide} Wide\&Deep consists of two parts: its deep model is the same as DNN, and its wide model is a linear model.
		\item \textbf{PNN} \cite{qu2016product} PNN uses a product layer to capture interactive patterns between interfield categories.
		\item \textbf{DIN} \cite{zhou2018deep} DIN uses the mechanism of attention to activate related user behaviors, which can be regarded as an attentive Pooling.
		\item  \textbf{DIEN} \cite{zhou2019deep} DIEN uses GRU with attentional update gate to model the user interest pattern.
	\end{itemize}
	
	\subsubsection{Metrics}
	In CTR prediction field, AUC is a widely used metric \cite{fawcett2006introduction}. 
	It measures the items of order by ranking all these with predicted CTR, including intra-user and inter-user orders. 
	A variation of user weighted AUC is introduced in \cite{zhu2017optimized,he2016ups} which measures the items of intra-user order by averaging AUC over users and is shown to be more relevant to online performance on CTR prediction. 
	We adopt this metric in our experiments. 
	For simplicity, we still refer to it as AUC. 
	Although there are other metrics are widely used in recommender system such as MRR@k or Recall@k, but CTR task with AUC metrics is our prior concern,  i.e., $k=1$. Since in industrial video recommendation, there are some key positions that requires high CTR, for example, the first recommended video take over most of the user attention and straight directly impact user retention.
	The metric is calculated as follows:
	$$
	AUC = \frac{\sum_{i=1}^n \#impression_i \times AUC_i}{\sum_{i=1}^n \#impression_i}
	$$
	where $n$ is the number of users, $\#impression_i$ and $AUC_i$ are the number of impressions and AUC corresponding to the $i$-th user.
	Besides, $RelaImpr$ metric is used to measure relative improvement over models, and $RelaImpr$ is defined as below:
	$$
	RelaImpr = (\frac{AUC(measured~model)}{AUC(base~model)}-1) \times100\%
	$$
	
	\subsubsection{Experiment Settings}
	The embedding size of video and category are both 18, which then been concated as an embedding of 36. The dimension of user profile embedding is 36. Mapping $g$ is a linear transformation with transparent matrix of size $36\times36$, and mapping $\phi$ is a two  fully connect layers with size 72 and 36. The dimension of FC in guide loss is set to 18.
	$\lambda$ is set to 0.5. The Runge-Kutta methods is used as ODE solver. We follow the setting of BaseModels as their suggest. We train DTS on a GTX 1080ti for 5 epochs, with batch size set to 128. 
	
	\subsection{Results on Public Dataset}
	Remind that \textit{Deep Time-Stream framwork = BaseModel + Time-Stream Module}. 
	Thus, the effectiveness of the improvement brought by the Time-Stream Module should be verified.
	\begin{table}[t]                                
		\centering                            
		\caption{Model Comparison with AUC on Amazon Dataset. }  
		\begin{tabular}{cccc}                            
			\hline                            
			&    BaseModel    & DTS   &    RelaImpr    \\
			\hline                          
			DNN     &    0.7686    &    0.7789    &    1.34\%    \\
			PNN     &    0.7799    &    0.8304    &    6.48\%    \\
			Wide\&Deep     &    0.7735    &    0.8390    &    8.47\%    \\
			DIN     &    0.7880    &    0.8508    &    7.97\%    \\
			DIEN     &    0.8453    &    0.8981    &    6.25\%    \\
			\hline                          
		\end{tabular}                            
		\label{tab:Books}                        
	\end{table}    

	As shown in Table~\ref{tab:Books}, there exists some facts that: 
	(1) our proposed DTS clearly outperforms all the raw model on five BaseModel, which confirms the capacity of our model in learning impact of time. 
	(2) for the BaseModels, PNN could capture interactive patterns between inter-field categories, which beats DNN and Wide\&Deep.
	However, above three BaseModels use average pooling to compress user features into a fixed-length vector, which brings a difficulty to capture user's diverse interests effectively from rich historical behaviors. Moreover, 
	DTS could generate the better representation of user behaviors by considering time-stream information, and achieves improvements up to 8.47\%, which beats the performance of raw DIN. 
	(3) DIN with Time-Stream Module outperforms DIEN. DIN leverages attention mechanism and improves the expressive ability of the model greatly. 
	The follow-up work DIEN based on DIN further tries to capture the interest evolving process. Compared with DIEN, DIN with Time-Stream Module considers continuous time-stream information, and it could help CTR model to learn more powerful user representation compared with previous works.
	
	
	\subsection{Results on Industrial Dataset}
	We further conduct experiments on the dataset of the real short video platform.
	In practice, the max length of history behaviors is set as 100.
	\begin{table}[htbp]                                
		\centering                            
		\caption{Model Comparison with AUC on Industrial Dataset.}
		\begin{tabular}{cccc}                            
			\hline                             
			&    BaseModel    &    DTS    &    RelaImpr    \\
			\hline                             
			DNN     &    0.6385    &    0.6628    &3.81\%    \\
			PNN     &    0.6601    &    0.6763    &    2.45\%    \\
			Wide\&Deep    &    0.6478    &    0.7010    &    8.21\%    \\
			DIN     &    0.7008    &    0.7268    &    3.72\%    \\
			DIEN    &    0.7023    &    0.7412    &    5.54\%    \\
			\hline                             
		\end{tabular}%
		\label{tab:addlabel}%
	\end{table}%
	
	\begin{table*}[ht]                                                          
		\centering                                                            
		\caption{Ablation studies with AUC on the industrial dataset. ``w'' is the short for ``with'' and ``w/o'' is the short for ``without''. }                                                        
		\begin{tabular}{cccccc}                                                            
		\hline                                                         
			&    BaseModel  &    w/o adaptive step (RNN)    &    w simple form    &    w/o    guide loss    &    DTS      \\
			\hline                                                            
			DNN     &    0.6385    &    0.6532    &    0.6389    &    0.6441&    \textbf{0.6628}        \\                
			PNN    &    0.6601    &    0.6703    &    0.6721    &    \textbf{0.7095} &    0.6763       \\                
			Wide\&Deep    &    0.6478  &    0.6948    &    0.6802    &    0.7007 &    \textbf{0.701}      \\                
			DIN    &    0.7008   &    0.7002    &    0.7012    &    0.7096  &    \textbf{0.7268}       \\                
			DIEN    &    0.7023    &    0.7021    &    0.7045    &    0.7116 &    \textbf{0.7412}       \\    
			\hline                                                            
		\end{tabular}%
		\label{tab:application}%
	\end{table*}%
	As shown in Table \ref{tab:addlabel}, our Time-Stream framework could improve all of the BasedModels. 
	The BaseModels of DNN, PNN, and Wide\&Deep are widely used in industry and build on large scale distributed system, and the change of model would make a great effort. Our DTS could easily apply on these model with no changes made to the original architecture, and it brings at least 3\% improvement. It suggests that our DTS has great value of practical application.
	Similar to Amazon Dataset, DIN with Time-Stream Module outperforms raw DIEN, which confirms the capacity of our model.

	\subsection{Model analysis}
	In this subsection, we will show the effect of adaptive step, function form and guide loss, respectively.
	
	\subsubsection{Effect of adaptive step}
	To verify whether adaptive step helps to construct better representation, we conduct ablation study on fixing the step of hidden dynamics to demonstrate the effectiveness of adaptive step. 
	From Table~\ref{tab:application}, without adaptive step would perform worse than origin. The DTS with fixed step is equivalent with RNNs, when $\Delta t_i\equiv t_{i+1}-t_i, i=1,2,...,N$ are all constants. The time interval are not considered by RNN. Thus, compared with fixed step, adaptive step could evaluate the step of $f$ whenever neccessary.
	Besides, it could handle incorporate data which arrives at arbitrary time. 
	 Hence, our DTS could learn more accuracy users' interests, such that achieving better performance.
	
	\subsubsection{Effect of function form}
	When we use simple form of function $f$ discussed in General Framework, the performance is better than BaseModel as shown in Table \ref{tab:application}. 
	However, the improvement is still limited compares with the complex form that has more powerful expression.

	\subsubsection{Effect of guide loss}
	Moreover, we further explore the effect of guide loss. It uses non-click items as negative instances for enhancing discrimination. 
	As shown in Table~\ref{tab:application}, guide loss brings great improvements most of all BaseModels, which reflects the importance of supervision information when learning the representation of latent user interest. 
	
    
	\section{Related Work}
	\label{section:Related}
    By virtue of the strong ability of deep learning on feature presentation and combination recent CTR models transform from traditional linear or nonlinear models\cite{friedman2001greedy,rendle2010factorization} to deep models. Most deep models follow the basic paradigm of Embedding, Pooling and Multi-layer Perceptron (MLP) \cite{covington2016deep}. Based on this paradigm, many models pay attention to the interaction between features: Wide\&Deep \cite{cheng2016wide} combines low-order and high-order features to improve the power of expression; PNN \cite{qu2016product} proposes a product layer to capture interactive patterns between interfiled categories.
	DIN \cite{zhou2018deep} introduces the mechanism of attention to activate the user historical behaviors w.r.t. given target item locally, and captures the diversity characteristic of user interests successfully.
	
	Beyond that, several methods \cite{wu2017recurrent,hidasi2015session,yuan2019simple} are proposed for capturing dynamic information, since user behaviors are usually dynamic instead isolated. The dynamic information also could be regarded as a kind of context information in recommendation system, which is  distinguishable from features describing the underlying activity undertaken by the user within the context~\cite{dourish2004we}. Moreover, the sequential models usually get the better performance on capturing dynamic information.
	These works regard user-item interactions as a sequence and try to represent sequential user behaviors.
	\cite{yu2016dynamic} uses the structure of the recurrent neural network (RNN) to investigate the dynamic representation of each user and the global sequential behaviors of item purchase history.
	Some methods use RNNs to capture dynamic information in recommendation system, e.g. RRN~\cite{wu2017recurrent}, GRU4REC~\cite{hidasi2015session}, NextItNet~\cite{yuan2019simple}.
	\cite{zhou2018atrank} uses an attention-based sequential framework to model heterogeneous behaviors. 
	DIEN \cite{zhou2019deep} leverages GRU and designs an attentional update gate (AUGRU) to model the dependency between behaviors .
	Although improving the performance compared to non-sequential approaches, these RNN based methods still are no enough to represent the user interest evolution without considering time-stream information.


	Recent years, some studies \cite{chen2018neural,weinan2019mean} go further to explore the possibility of producing nonlinear functions using continuous ODEs, pushing the discrete approach to an infinitesimal limit.
	They introduce the numerical differential equations to the design of deep neural network.
	\cite{weinan2019mean} shows many effective networks, such as
	ResNet, PolyNet, FractalNet and RevNet, can be interpreted as different numerical discretizations of differential equations. 
	Compared with deep neural networks, there are several advantages with a continuous approach, including to flexible choose the number of evaluations on recurrent networks,
	well-studied and computationally-cheap numerical ODE solvers. 
	However, to the best of our knowledge, there not exists previous works that leveraging ODEs to represent user interest evolution on CTR model.

	\section{Conclusions}
	\label{section:conclusions}
	In this paper we propose a novel Time-Stream framework, that adaptively mines the users‘ continuously evolving interests from rich historical behavior, by leveraging neural ODE that parameterizes the derivative of the hidden state using a neural network.
	Unlike recurrent neural networks, which require discretizing observation and emission intervals, continuously-defined dynamics can naturally incorporate data which arrives at arbitrary time. 
	We also propose guide loss to control the error of ODE solver. 
	Extensive experiments show that our model can generate a more precise user feature at an arbitrary time.
	\section{Acknowledgments}
    This research was partially supported by National Key Research and Development Program (2017YFB1001903) and Alibaba Group through Alibaba Innovative Research Program.
	\bibliography{references.bib}
	\bibliographystyle{aaai}
\end{document}